# Lasso–Ridge-based XGBoost and Deep_LSTM Help Tennis Players Perform better

WANKANG, (W.K.), ZHAI[#]

Houston International Institute, Dalian Maritime University, Dalian, Dalian 116026, China, zhai@dlmu.edu.cn

YUHAN, (Y.H.), WANG[#*]

Houston International Institute, Dalian Maritime University, Dalian, Dalian 116026, China, 557537577mm@dlmu.edu.cn

## Abstract

  Understanding the dynamics of momentum and game fluctuation in tennis matches is crucial for predicting match outcomes and enhancing player performance. In this study, we present a comprehensive analysis of these factors using a dataset from the 2023 Wimbledon final. Initially, we develop a sliding-window-based scoring model to assess player performance, accounting for the influence of serving dominance through a serve decay factor. Additionally, we introduce a novel approach, Lasso-Ridge-based XGBoost, to quantify momentum effects, leveraging the predictive power of XGBoost while mitigating overfitting through regularization. Through experimentation, we achieve an accuracy of 94% in predicting match outcomes, identifying key factors influencing winning rates. Subsequently, we propose a Derivative of the winning rate algorithm to quantify game fluctuation, employing an LSTM_Deep model to predict fluctuation scores. Our model effectively captures temporal correlations in momentum features, yielding mean squared errors ranging from 0.036 to 0.064. Furthermore, we explore meta-learning using MAML to transfer our model to predict outcomes in ping-pong matches, though results indicate a comparative performance decline. Our findings provide valuable insights into momentum dynamics and game fluctuation, offering implications for sports analytics and player training strategies.

**Keywords:** Parallel Model; XGBoost; MAML; LSTM; Tabular Data

## 1. Introduction

### 1.1 Problem Background

  In recent years, tennis has been loved and sought after by many people, and more and more people want to pay attention to tennis matches. The 2023 Wimbledon men's final was particularly high-profile, as Carlos Alcaraz defeated Novak Djokovic to take the men's title.Moreover, he won because of "momentum" and "momentum."

  In dynamic contests, strategic momentum and psychological momentum potentially co-exist, which makes it difficult to distinguish between the two. We employ the setting of professional tennis, which allows us to separate psychological from strategic momentum. In tennis, converting a break point —momentum fluctuations are called turning points— potentially triggers both strategic momentum—due to a change in the relative position of the players—and psychological momentum—due to a change in the perception of the players. Players can control the momentum in the game by understanding its role and adopting appropriate strategies.

---

[#]These authors contribute equally to this work
\* Corresponding author.

## 1.2 Literature Review

With the development of artificial intelligence technology, the research and application of deep learning in the field of sports are also gradually increasing. Applying artificial intelligence technology to competitive sports tactics is not only a hot spot of theoretical research at home and abroad, but also an effective means to promote the high-quality development of competitive sports. The potential application of deep learning in sports analysis, exploring the potential of deep learning as a tool for analyzing sports tactics. At present, in the aspects of player and ball tracking, action behavior analysis, and performance evaluation, the analysis method based on deep learning has shown good performance. It has excellent potential for development in the future.

The research results of deep learning-based video analysis of competitive sports performance by scholars such as Gong Guoyu strongly prove the application potential of deep learning technology in sports and its revolutionary impact[1]. In the analysis of tennis technical action, it is necessary to use many new technologies to complete, of which artificial intelligence technology is one. Liu Mengxin and other scholars concluded that the application of artificial intelligence in tennis technical action analysis provides more accurate and personalized training guidance for athletes and coaches through many applications of artificial intelligence in tennis technical action analysis. Through deep learning and computer vision technology, artificial intelligence can automatically identify and analyze an athlete's movements, helping athletes improve hitting form, stride, and power control[2]. Shi Guangbinetal. proposed a small-scale automated tennis tactical assistant guidance system based on the tactical analysis and guidance of multi-shot deadlock rounds. They built a decision-making model of optimal return drop point under multi-shot deadlock. It also shows that machine learning is beneficial[3].

## 1.3 Our Work

Considering the background and the problems, our work mainly includes the following:
● We first selected the men's singles final at Wimbledon as our dataset, and then completed a visualization operation based on the scores of two players, and identified which time period the players performed most prominently. At the same time, the impact of serving rounds on the game progress and results was considered.
●We use the Fibonacci sequence to quantify momentum and update the dataset by obtaining the changes in athlete momentum compared to the previous moment. Afterwards, we use XGBoost ensemble learning to model and use different permutations and combinations of momentum features to determine the impact of momentum on the competition.
● Afterwards, we used LSTM deep learning to model and calculated the gradient of momentum over time. We then visualized the results to analyze the fluctuations of the game and determine when the game process would shift from leaning towards one player to leaning towards another.
● Finally, we use meta learning MAML to complete transfer learning for other matches in

Wimbledon. We use datasets from multiple Wimbledon matches as inputs to train the meta learning deep learning network, and then use the remaining matches as test sets to measure the accuracy of the model. Afterwards, we applied the model to table tennis matches and used the XGBoost model to measure the degree of influence of momentum in table tennis matches.

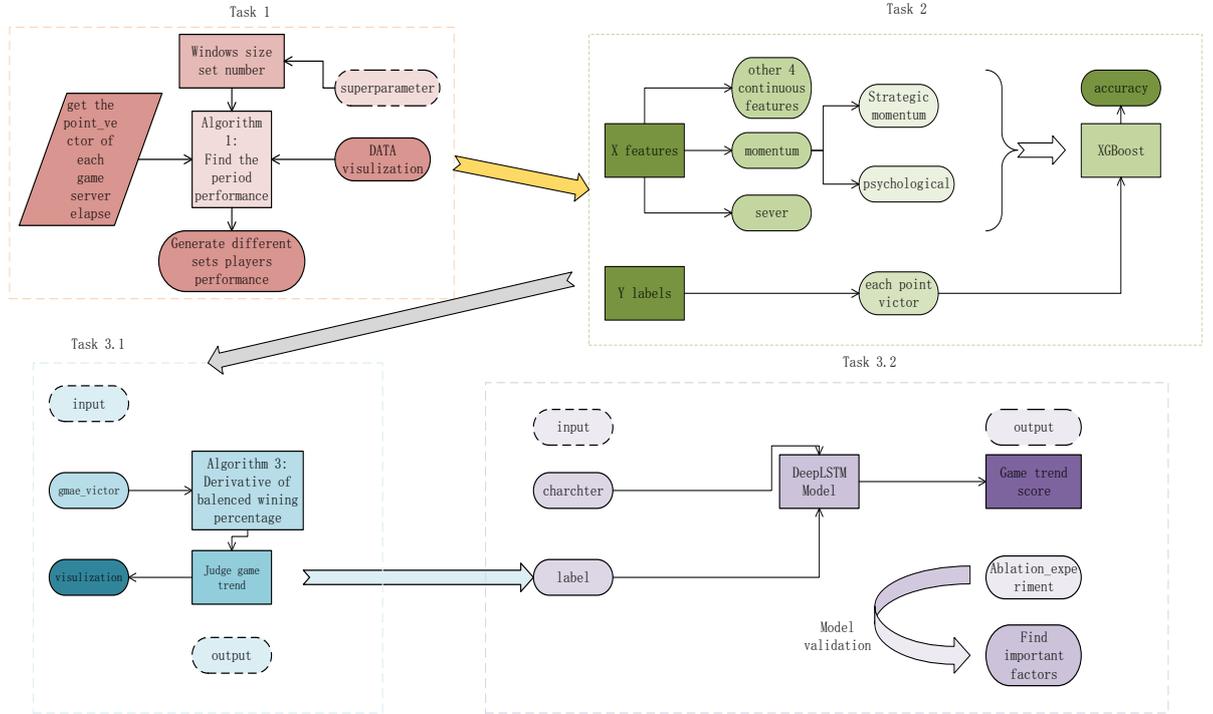

Figure 1     Flow chart of our work

## 2 Notations

The key mathematical notations used in this paper are listed in Table 1.

Table 1    Notations used in this paper

| Symbols | Description |
| --- | --- |
| $\alpha$ | The ratio of Lasso and Ridge loss |
| $\beta$ | The decay rate of serving |
| $w_v$ | The window size of game victor |
| $w_s$ | The window size of server victor |
| $s$ | Server for the whole time |
| $v$ | Point Victor for the whole time |
| $t$ | The elapsed_time of each points |

## 3  Player performance time period analysis

The first chapter is the analysis of the players' performance time

In the first chapter, we successfully built a model to judge player's performance. (WINJUD) and visualize it through the window. Through our membrane type, we analyze the match's result to evaluate the player's performance in each period and draw it into a heat map to show the difference between the two players in the competition process.

## 3.1 How WINJUD predicts

The most direct indicator of competitive performance in a game is when a player wins every game in a period. And whether it succeeds or not, it directly relates to the server. Thus, in analyzing the state of play, our input is the probability that the player will win a match in each five-game period and the likelihood that the player will serve in those five games. And add a serve factor.

Our algorithm predicts player performance according to the following Algorithm1.

**Algorithm 1: WINJUD**
**require:** $v$ : Point Victor for the whole time
**require:** $s$ : Server for the whole time
**Super parameters:** $\beta$ : The decay rate of serving
**Super parameters:** $w_v, w_s$ : Step size hyperparameters
1: **for all** $i\ in\ range(v)$ **do**
2:  **if** $i >= w_v$ **do**
3:   **current_victor_window** is the window from $i - w_v$ to $i$
4:   **Do** step 3 for $w_s\ to\ find$ **current_server_window**
5:   **compute** each player's WINJUD score:
6:   **player 1's score** :
7:    count the number of 1 in **current_victor_window** + $\beta *$ count the number of 2 in **current_server_window**
8:   Do the same for **player 2's score**
9: **end for**

## 3.2  1702 Wimbledon final match analysis

In this algorithm, we add a sliding window to calculate each player's performance over time. In Figure 1, we plotted the heat maps of each player during the game and compared them.

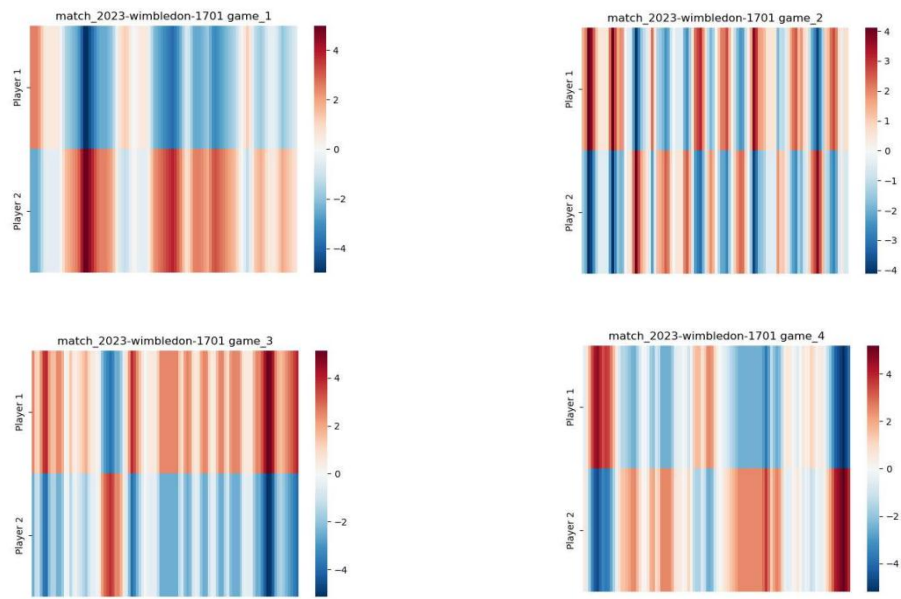

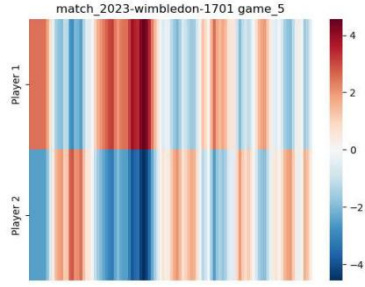

Figure 2  Player's WINJUD score

Figure 2 shows that Player 1 is Carlos Alcaraz (CA), and Player 2 is Novak Djokovic (ND). At the beginning of the match, ND takes the lead and successfully wins the first match. In the second game, both sides were very anxious, playing You to Me, and everyone had a good moment, but in the end, CA won the game with the final sprint. In the third game, ND performed well in the early stage but eventually served worse and worse and finally lost. The initial stage was not smooth in the fourth game, but it finally broke out and won. The last game, by a brief good performance, ultimately did not continue and lost the finals.

We analyzed the corresponding moment of the best performance in the chart and obtained Table 2. Therefore, before and after Table 1, the players' performance was awe-inspiring!

Table 2  Performance Best Time

|  | game_1 | game_2 | game_3 | game_4 | game_5 |
| --- | --- | --- | --- | --- | --- |
| Carlos Alcaraz | 0:08:02 | 0:52:49 | 2:19:40 | 3:52:02 | 4:05:45 |
| Novak Djokovic | 0:01:19 | 0:38:34 | 2:52:06 | 3:09:09 | 4:18:08 |

# 4  Momentum in the tennis match

In general knowledge, momentum is an indicator that directly reflects the status of a player. It can evaluate whether a player tends to win or lose within a certain period of time, whether he is going to counterattack or is always suppressed. In a basketball game, if a team scores consecutive points and hits a climax, the coach of the opposing team will usually call a timeout in time to control the situation. Momentum is equally crucial in tennis, especially considering that the service round plays a decisive role in the score. How players adjust their game status and game plan according to changes in momentum on the court is a difficult problem that the coaching staff needs to overcome.

Of course, some coaches are skeptical about the impact of momentum on the game. He believes that momentum will not directly lead to a one-sided trend in the game. Below, our team first gives a reasonable quantification method for momentum, and then uses the XGBoost model to analyze the accuracy of the prediction results, aiming to analyze whether this view is reasonable.

Momentum can be divided into strategic momentum and psychological momentum. Psychological momentum refers to the psychological changes that occur after a player wins or loses a small game. On the other hand, a player's mistakes can also have a significant impact on psychological momentum. In tennis matches, if a player throws an ACE ball, their psychological momentum is bound to increase significantly. On the contrary, if a player makes unforced mistakes or makes two consecutive mistakes, their psychological momentum will be greatly weakened.

For strategic momentum, it is more based on the level of competition results. If the player is currently leading by a large score, it can be said that the player has a significant strategic momentum and has the momentum to win the game with one go. Similarly, if the player is lagging behind in a large score, then the player has no strategic momentum or negative strategic momentum, indicating that the final outcome of the game is not optimistic for the player.

## 4.1 Strategic Momentum

In tennis matches, strategic momentum only needs to consider two factors: the win or loss of each game and the win or loss of each set. In addition, we also take into account the different effects of winning streaks and losing streaks on strategic momentum. We believe that the addition of multiple consecutive wins to strategic momentum is significantly stronger than the impact of a single victory on momentum, so we consider power operations to calculate the momentum of consecutive wins or losses. Here, we use the performance of player1 as the standard for measuring momentum. In the provided data, we mainly focus on the four columns of "p1_sets", "p2_sets", "p1ugames", and "p2_games" to calculate the strategic momentum. The specific algorithm is presented in Algorithm 2.

---
Algorithm 2: Calculating the Strategic momentum

**Notation:**
    momentum1: Based on different sets
    momentum2: Based on different games
    $b_0$:   Value of the first row
    $m_1$:   Values of "p1_sets" or "p1_games" under given index
    $m_2$:   Values of "p2_sets" or "p2_games" under given index
    n:   an integer

Calculating momentum1:
1:   Grouped the column by "set_no"
2:   Set   $b_0 = 1$   to initialize
3:   *for* i = 1 to len(p1_sets) *do*
4:      b = $m_1$ - $m_2$
5:      *if* b == n and b > 0 *then*
6:         momentum1 = momentum1 * (1.5 ^ n)
7:      *elif*   b == n and b< 0 *then*

| | |
|---|---|
| 8: | momentum1 = momentum1 / (1.5 ^ n) |

Calculating momentum2:

| | |
|---|---|
| 1: | Grouped the column by "set_no" and then grouped by "game_no" |
| 2: | Set $b_0 = 5$ to initialize |
| 3: | **for** i = 1 to len(p1_games) **do** |
| 4: | $b = m_1 - m_2$ |
| 5: | **if** b == n and b > 0 **then** |
| 6: | momentum2 = momentum2 * (1.2 ^ n) |
| 7: | **elif** b == n and b < 0 **then** |
| 8: | momentum2 = momentum2 / (1.2 ^ n) |

Calculate Final Strategic momentum:

| | |
|---|---|
| 1: | momentum_strategic = momentum1 * momentum2 |

## 4.2 Psychological momentum

For psychological momentum, there are relatively many factors that affect its size. We considered whether a player was able to serve an ace, whether he committed an unforced error, whether he committed a double fault, and whether the player was on a winning streak or losing streak in the previous possession. For each influencing factor, after extensive experiments, we assigned them the most reasonable parameters so that the calculated psychological momentum can better reflect the player's game momentum at that moment.

In the same way, in the process of winning every small point, winning streak or losing streak also plays a crucial role in the psychological momentum of the players. Therefore, we also considered that multiple winning streaks or losing streaks will have a greater bonus or impact on players. When calculating psychological momentum, we consider using the ***Fibonacci sequence*** to measure and calculate a player's psychological momentum. The specific algorithm will be presented in Algorithm 3.

Algorithm 3: Calculating Psychological momentum

**Notation:**

victor_number: The player number that win one point in a single game
$count_1$: Number of 1s that occur before specific moment
$count_2$: Number of 2s that occur before specific moment

| | |
|---|---|
| 1: | Grouped the column by "set_no" and then grouped by "game_no" |
| 2: | Initialize the momentum of first row of each game to be 1 |
| 3: | **for** i =1 to len(point_victor) **do** |
| 4: | if victor_number is 1 then |
| 5: | $count_1$ + 1 and $count_2$ = 0 |
| 6: | **else** |
| 7: | $count_2$ + 1 and $count_1$ = 0 |
| 8: | Update momentum based on $count_1$ and $count_2$ using ***Fibonacci sequence*** |

Here, we first select a match (match 1701, Carlos Alcaraz vs. Novak Djokovic) as an example to quantitatively analyze the relationship between athletes' psychological momentum and match time (in seconds). We uniformly select player 1 as an example to measure, and then visualize the results（without inserting continuous values). The visualized results are shown in Figure 3. From the graph we can see that in most of the time, the psychological momentum of player 1 is a little bit above 0 which stands that the two players are **nip and tuck.**

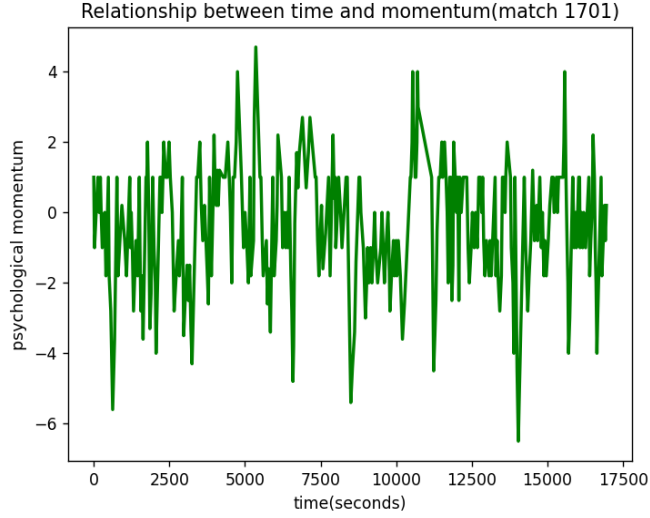

Figure 3　The psychological momentum at different time

In order to prepare for machine learning and deep learning later, we need to select features for the provided data set. During the modeling process, we selected a total of 6 features to put into the subsequent XGBoost model, namely the two momentums just calculated and the serving rounds in the given data, the running distance of the two players, and the hitting Number of hits and ball speed.

### 4.4　XGBoost Model

XGBoost is a classic ensemble boosting algorithm framework that builds a series of decision trees and combines their predictions to create a more powerful and accurate model. Its essence is to integrate many basic models together to form a strong model.

First, according to the needs of the XGBoost model, the most important thing is to construct the objective function. Assume that there is a data set consisting of n samples and m features at this time, and its initial objective function is as shown in (1).

$$\Psi^{(t)} = \sum_{i=1}^{n} \ell(y_i, \hat{y}_i^{(t)}) + \Omega(f_t) \tag{1}$$

where $y_i^{(t)}$ is based on the predicted value of the first nth trees. The loss function is given by (2).

$$\ell(y_i, \hat{y}_i^{(t)}) = \frac{1}{n}\sum_{i=1}^{t} w_i(y_i - \hat{y}_i)^2 \qquad (2)$$

$\Omega(f_t)$ is the complexity of the model and it is given by (3).

$$\Omega(f_t) = \lambda_1 \rho \|w\|_1 + \frac{\lambda(1-\rho)}{2}\|w\|_2^2 \qquad (3)$$

From that we can get our cost function to be like as (4).

$$\text{cost}^{(t)} = \arg\min(\sum_{i=1}^{n}\frac{1}{n}\sum_{i=1}^{t} w_i(y_i - \hat{y}_i)^2 + \lambda\rho\|w\|_1 + \frac{\lambda(1-\rho)}{2}\|w\|_2^2) \qquad (4)$$

For the loss function $\ell(y_i, \hat{y}_i^{(t)})$, it involves many models (decision trees) participating together and is obtained through additive training. After training the first tree, if the first tree is not trained well, use the second tree to continue training, and so on, train the K-1th tree, and finally train the Kth tree. When training the Kth tree, the previous first tree to the K-1th tree are all known, and only the Kth tree is unknown. Therefore, based on the previously constructed decision tree known, the Kth tree is constructed tree. The logical structure is shown in Figure 4.

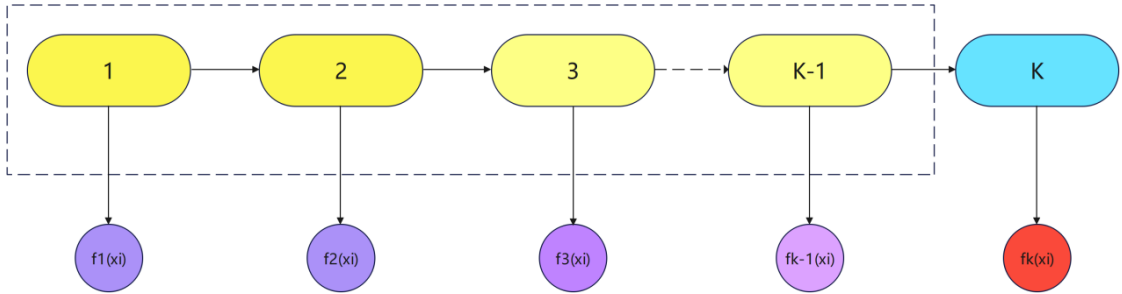

Figure 4　　The basic logic of the Boosting

Then we tried to solve the optimization problem of the objective function. In the original text of XGBoost, Chen[5] used the Taylor series expansion method to approximate the objective function. Expand the objective function into a polynomial. The higher the polynomial order, the higher the approximation to the objective function.

According to the principle of Boosting, we can get the relationships between the (K-1)th tree and the Kth tree, which is shown in (5).

$$y_i^{(k)} = y_i^{(k-1)} + f_k(x_i) \qquad (5)$$

Then after using Taylor series, we can get our final loss function in (6).

$$L(t) = \sum_{i=1}^{n}[g_i f_t(x_i) + \frac{1}{2}h_i f_t^2(x_i)] + \Omega f(t) \quad (6)$$

## 4.5 Lasso - Ridge based XGBoost

Now that we have obtained the feature data and sample data that need to be selected from the data of a game, we will put the processed data into our XGBoost elastic net model for model training and result prediction.

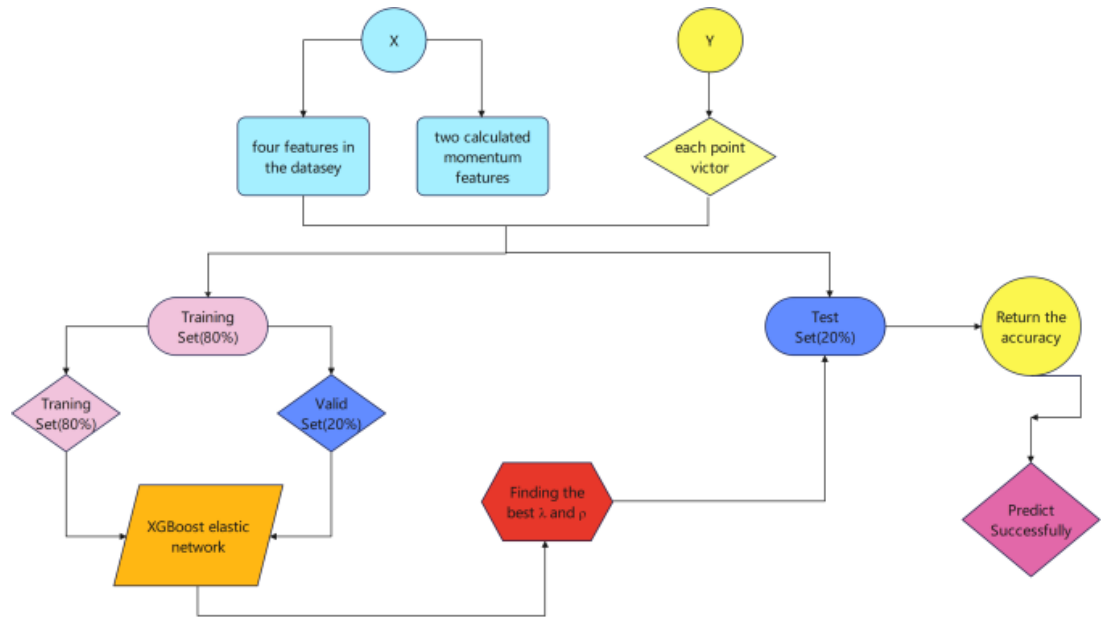

Figure 5　The flow chart of processing the Lasso - Ridge based

Figure 5 shows the whole work of our team on the processing the XGBoost network. We use the conventional method of dividing the data set, that is, 80% is used as the training set and 20% is used as the test set to train the model. After that, 80% of the training set is further divided into 80% of the training set and 20% of the verification set and input into the XGBoost network for parameter training, with the intention of finding the best $\lambda$ and $\rho$.

The trained parameters are then applied to the test set and a correct rate is returned. The correct rate of prediction can reflect the prediction accuracy of the model and also reflect the impact of momentum on the game. The meaning of the accuracy rate is: first use XGBoost to predict the probability of player 1 winning the game, then calculate the probability of player 1 winning the game in the actual game, and then compare the two to get a prediction accuracy rate, which is the final The output results can reflect not only the excellence of the model, but also the impact of momentum on the game.

We used the method of controlling variables to calculate the probability of successful model prediction in four cases: "no momentum added", "only strategic momentum added",

"psychological momentum only added" and "both momentum added". We selected multiple Wimbledon match data to predict the output of four probabilities, and the final results are shown in Table 3.

Table 3  Prediction accuracy of the XGBoost Model of selected

| Match_Id | Only psychological momentum | Only strategic momentum | Not Use | Use Both |
| --- | --- | --- | --- | --- |
| 1302 | **0.8536585** | 0.8292683 | 0.8123654 | 0.780489 |
| 1304 | **0.9705882** | 0.8970588 | 0.9411765 | 0.897059 |
| 1314 | **0.9152354** | 0.8108108 | 0.7981542 | 0.837838 |
| 1401 | **0.8444444** | 0.8000000 | 0.8222222 | **0.844444** |
| 1701 | 0.7313433 | **0.8059701** | 0.7164179 | 0.776119 |

### 4.6  Model evaluation and the analysis of momentum

Through the analysis of the data in Table 3, we can see that for most games, only considering psychological momentum has a greater impact on the game trend, and the prediction results are more accurate. When the impact of any momentum on the game is not taken into account, the results are not ideal. And if both strategic momentum and psychological momentum are taken into account, the results of some game predictions are not as accurate as if only psychological momentum is considered. In particular, for the final of Wimbledon (1701), the results predicted by our model found that only considering strategic momentum had a greater impact on the game, while the impact of psychological momentum on the game was not significant. One possible explanation is that for two of the world's best players, if they can get a score lead in a tense game, it will have a direct impact on the game.

Therefore, we can conclude that momentum plays a decisive role in the game. In most games, psychological momentum plays the most significant role. But if two athletes are of similar quality, strategic momentum can be a big factor in dominating the game.

## 5  Predicting Match Fluctuations Using Deep Learning

### 5.1.1  Match Fluctuation Score DBWP Principle

In this chapter, we define an algorithm derivative of balanced winning percentage (DBWP) to generate match fluctuations and visualize the results. We utilize our DBWP scores as labels for match fluctuation prediction using the Deep_LSTM model we built. We can predict most of the match trends and fluctuations. For this regression problem, the results show that the average error of our results is around 0.03-0.04, and our model is able to converge to fit. More critically, we have successfully predicted the factors affecting match fluctuations through ablation experiments.

**Algorithm 4: DBWP**
---
**require:** $v$ : Point Victor for the whole time
**require:** $t$ : The elapsed_time of each points
**Super parameters:** $w_v$ : Step size hyperparameters
1: **for all** *Point Victor* $v$ **do**
2:     ***find*** Before and after $w_v$ ***games victor***, ***Do***:
3:         ***Calculate*** *the ratio of* $Player_{1(2)}$ *winning rate*
4.         *Linearly interpolate the win rate results*
5:         ***Calculate*** *the* ***derivative*** *of each winning rate with respect to time.*
5:
*Only the derivative value of the corresponding* **elapsed_time** $t$ ***index*** *is retained*
    6. **end for**
---

With Algorithm 4, we calculate the inverse of the difference between the before and after winning percentage to determine the fluctuation of a game. Because it calculates the trend of the winning percentage before and after each moment, he can reflect the rapidly changing fluctuation situation on the field very well. We visualize and analyze it, and we can find that the fluctuation of the game is surprisingly closely related to the two kinds of momentum we defined in Task2!

### 5.1.2 Relationship between Momentum and Match Fluctuations

In Figure 6, we plot the DBWP score we calculated in 5.1.1 against Time, Psychology Momentum, Strategy Momentum, and we find that DBWP is correlated with Psychology and Strategy. In Algorithm 4, we calculate the inverse of the difference between the winning percentage before and after to determine the fluctuation of a game. Because it calculates the trend of the winning percentage moment, it can reflect the rapidly changing fluctuation situation in the field very well. We visualize and analyze it, and we can find that the fluctuation of the game is surprisingly closely related to the two kinds of momentum we defined in Task 2.

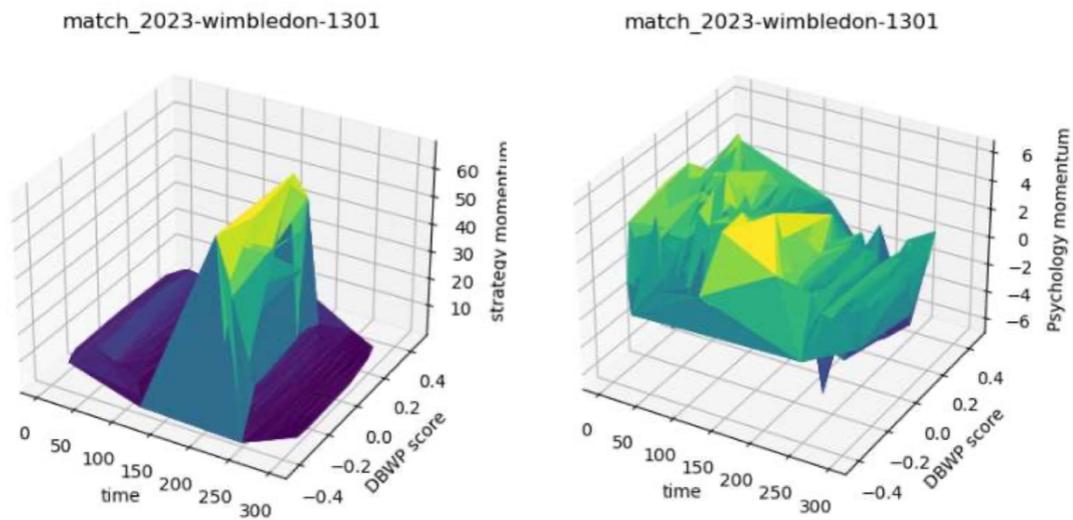

Figure 6　DBWP Psychology and Strategy

We can analyze from the 3D images that at each moment, DBWP and Momentum are strongly correlated. DBWP score at each moment We can analyze from the 3D images that at each moment, DBWP and Momentum are strongly correlated. DBWP score at each moment.

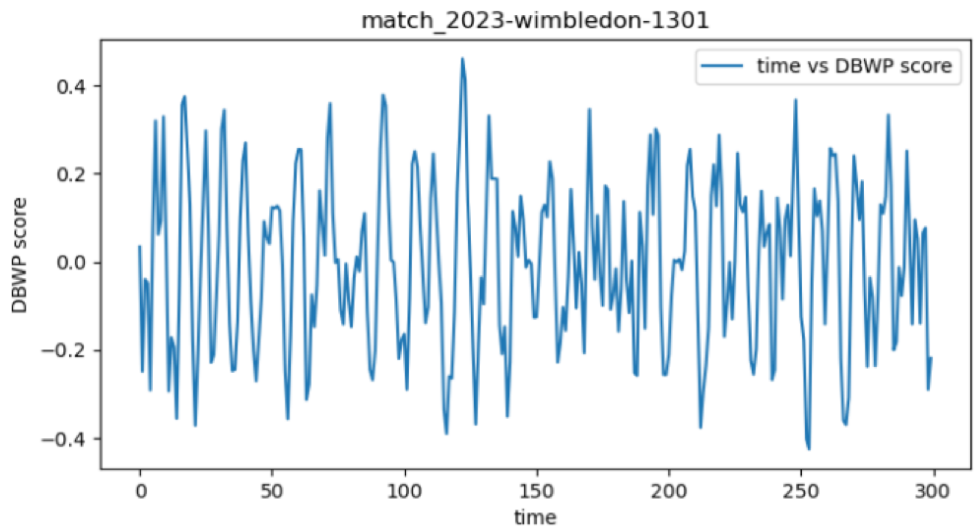

Figure 7　DBWP score in Match 1301

Figure 7 shows the predicted player 1 DBWP score as a function of time, and we find that the state of the game changes rapidly in the first game of 1301, but in general, Player1's performance is gradually declining.

### 5.1.3　Deep_LSTM model

In 5.1.3 we proposed the Deep_LSTM model. In Figure 5, the Deep_LSTM model is an advanced deep learning network model that consists of two networks in parallel and sums the results of the two networks at the end. The left side of it is a network that consists of advanced L2-Loss and Huber-Loss together. This network has advanced methods such as Dropout layer, SELU layer and Adam optimizer. The right layer of the network applies LSTM Long Short Term Memory. Traditional feed-forward neural networks assume that the inputs at moment t are unrelated to the inputs at moment t+1, but in many cases, this is not the case. For example, the momentum factor of our current tournament problem. Momentum at moment t and momentum at moment t+1 are closely related, so we recursively network to join the time series. However, it is clear to us that solving short time series using only RNNs will not reveal the long term dependency problem. We know that for strategy momentum, the relationship between set and set has a huge impact. So we introduced LSTM to our neural network. So we can learn high dimensional features while introducing the time variable.

Figure 4 shows the predicted player 1 DBWP score as a function of time, we find that in the first game of 1301, the state of the game is changing rapidly, but in general, Player1's performance is gradually declining.

## 5.2   Deep_LSTM model

In 5.2 we proposed the Deep_LSTM model. In Figure 5, the Deep_LSTM model is an advanced deep learning network model that consists of two networks in parallel and sums the results of the two networks at the end. The left side of it is a network that consists of advanced L2-Loss and Huber-Loss together. This network has advanced methods such as Dropout layer, SELU layer and Adam optimizer. The right layer of the network applies LSTM Long Short Term Memory. Traditional feed-forward neural networks assume that the inputs at moment t are unrelated to the inputs at moment t+1, but in many cases, this is not the case. For example, the momentum factor of our current tournament problem. Momentum at moment t and momentum at moment t+1 are closely related, so we recursively network to join the time series. However, it is clear to us that solving short time series using only RNNs will not reveal the long term dependency problem. We know that for strategy momentum, the relationship between set and set has a huge impact. So we introduced LSTM to our neural network. So we can learn high dimensional features while introducing the time variable.

$$\ell(y, f(x)) = \begin{cases} \frac{1}{2}(y - f(x))^2 + (y - f(x))^2 & if \ |(y - f(x)| \leq \delta \\ |(y - f(x))^2|\delta - \frac{1}{2}\delta^2 + (y - f(x))^2 & if \ |(y - f(x)| \leq \delta \end{cases}$$

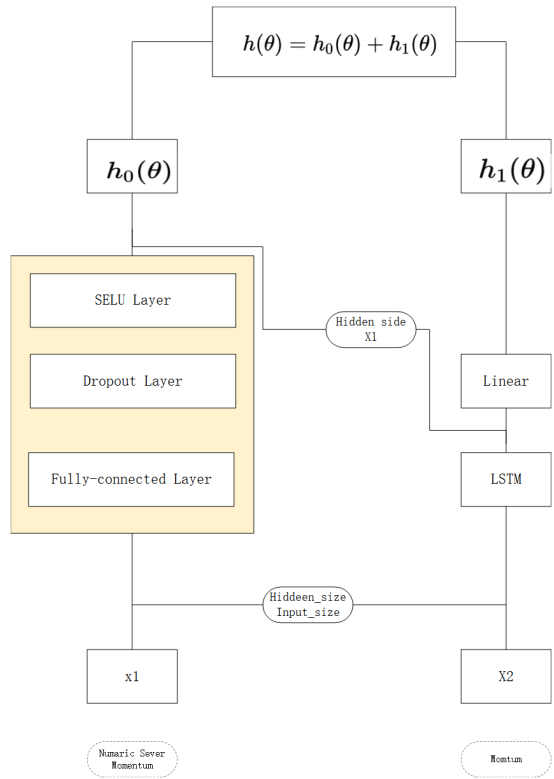

Figure 8　Deep_LSTM Baseline

For server, we consider that only considering 1's and 2's serves does not characterize it, so we change the prediction results to those for player1's fluctuations. We change server to 1 and -1. We consider that the result of the self-serve favors player1's mastery of the game, so we assign +1 to player1's free throws.

As a matter of fact, for server, we consider that only considering 1's and 2's serves does not characterize it, so we change the prediction result to the prediction result of player1's fluctuation. We change the server to 1 and -1. We consider that the result of the self-serve is in favor of player1's mastery of the game, so we assign +1 to the 1-penalty.

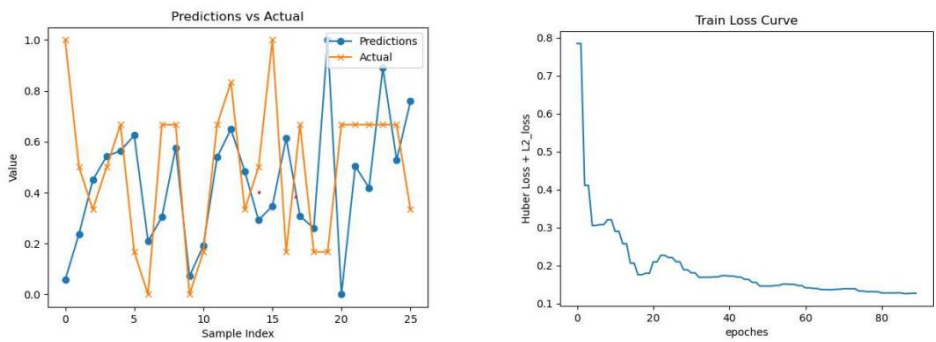

Figure 9　Model validation and loss curve

Figure 9 shows the difference between our predicted and true values, and the loss curve. We find that our model can be successfully fitted and eventually converge. We performed each experiment ten times and finally calculated the average value. Find the best factors, server and mind.

We intercepted the model evaluation data for the three games 1301 1302 1303 in the table. The full data is in the Appendix. Figure 6 shows the difference between our predicted and true values, and the loss curve. We find that our model can be successfully fitted and eventually converge.

We performed each experiment ten times and finally calculated the average value. Find the best factors, server and mind. Then we intercepted the model evaluation data for the three games 1301 1302 1303 in the Table 4. The full data can be found in the Appendix.

Table 4    The model evaluation data for three games

| For All | MSE-Mean | 0.03529891 | 0.0545164 | 0.05927794 |
|---|---|---|---|---|
| Lake of momentum | MSE-Mean | 0.03921655 | 0.0581546 | 0.06035726 |
| Lake of server | MSE-Mean | 0.03652814 | 0.0567297 | 0.06038466 |

Therefore, we recommend that coaches, when serving themselves, take control of the tempo! Because this is the key factor that affects the game. But the most important thing is to adjust a good mindset, we can see that momentum has the greatest impact on the result Therefore, we recommend that coaches, when serving themselves, to get the rhythm right! Because this is the key factor that affects the match. But the most important thing is to adjust a good state of mind, and we can see that momentum has the greatest impact on the result.

# 6 Model migration and multi-task adaptation

In this chapter, we first use the Model-Agnostic Meta-Learning (MAML) Model for meta-learning analysis of all Wimbledon matches. Then, we use the XGBoost model to predict the influence of momentum in table tennis matches.

## 6.1 Using meta-learning to optimize the model in multiple matches

### 6.1.1 MAML principle analysis

Meta-learning is a branch of deep learning in recent years, among which MAML is one of the algorithms with excellent performance under meta-learning. The model was proposed in 2017 by Professor Chelsea Finn of Stanford University and has since been refined by Professor Aravind Rajeswaran. Its avant-garde idea of a model-independent learning parameter distribution space is still impressive today. I stand based on the predecessor model, the tennis "momentum" of the problem of multiple-match data mining.

The original intention of meta-learning is to make the model find the best solution in a few samples. It still applies to this problem. Although our data is time series data, it is still challenging to find a suitable fitting model due to the heterogeneity and non-randomness of our data. Therefore, our team came up with the idea of using the MAML method for initial parameter learning. By learning a large number of matches, the reasonable parameters of the model are fitted under the stochastic timing problem. We give a large amount of prior knowledge to the model, continuously modify the initialization parameters through low epochs and high support, and finally find a reasonable hyperparameter so that the model can be fitted quickly.

MAML is mainly implemented by both internal and external cycles. The inner loop propagates forward for all tasks, updates to but does not update the model, and accumulates for each task through the outer loop.

### 6.1.2 MAML experimental

Our experiment was set up according to Figure 10: We took 1301-1316, 1401-1408, 1501-1504 as the support set to find the samples of the initial parameters. We made our final predictions for three races, 1,601,162,1701 as Query.

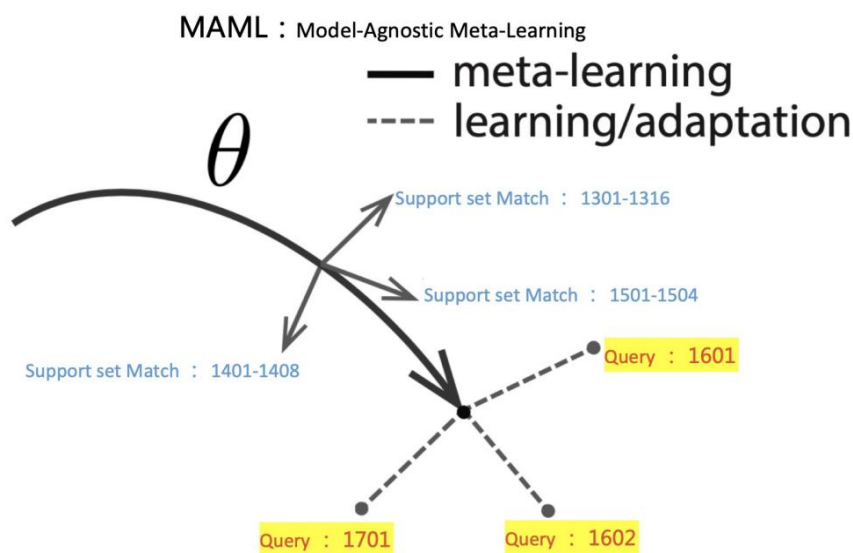

Figure 10　Maml updating parameters

### 6.1.3 Experiment settings

After initializing the model and optimization parameters through the support set, we Fine-tune the three queries to accommodate the new data distribution fully. Fine-tune's policy is the same as the model Deep_LSTM in Task 3. We use the Deep and LSTM layers in parallel to retrain the initialized hyperparameters by five epochs.

It is worth noting that we added the L2_loss function. Since the Huber we use is not sensitive to outliers, while the MSE loss function is susceptible to outliers, we believe no extreme outliers will occur in the inner process. MSE can effectively converge the internal network while the outer network uses the Huber loss function. It can avoid the interference of outliers and improve the robustness of the network.

Because meta-learning generalizes the network model by learning shallow features of multiple samples, we set only three epochs for the inner network to accumulate losses. The Table 5 shows the hyperparameter Settings for our meta-learning.

Table 5　Super Parameters for MAML

| Super Parameters | |
|---|---|
| mata_lr | 1.00E-04 |
| inner_lr | 1.00E-04 |
| input_dim | 240 |
| train_test | 0.8 |
| out_dim | 1 |
| tasks_per_batch | 4 |
| inner_loss_fn | MSELoss |
| Outer_loss_fn | HuberLoss |

### 6.1.4 Meta-learning outcome evaluation and analysis

We ran MSE and compared it to Task3, following the Settings in Task3. Our assessment of outcomes through meta-learning is as follows:

We found that the meta-learning results are close to the previous Deep_LSTM and analyzed why.

First, randomness tends to be heavy in tennis matches, even though we use what we know enough to predict changes in match fluctuations. But game fluctuations are not only numerical and sometimes do depend on luck. Any factor other than strength and statistics can sometimes affect the game's volatility. Think of the unevenness of the ground when a tennis ball falls, the slight change in the elasticity of the tennis ball, the breeze in the outdoor environment, and so on. Any one of these factors makes the race unknowable.

Secondly, in the 1602 game, the strength gap is enormous. Player 2 is Novak Djokovic, who has won the championship for many consecutive years. It is too subjective for us to predict the fluctuation of players in standard competitions only by using the scores and values of matches and other data, ignoring the strength factor of players themselves.

Finally, our model could be better. To fit the model better, we set only 0.05 parameter L2_Loss. We worry that if epochs are too high, the model will be overfitting, leading to a significant deviation in the prediction results. The loss of the outer Huber alone cannot adequately predict high-dimensional features. The loss function can be improved in the future.

# 7　Model Evaluation and Further Discussion

In this work, we use three models to mining data in tennis.
They are Lasso - Ridge based XGBoost, Deep_LSTM and MAML.
## 7.1 Strengths

Originality ：we thought of using the derivative of the difference in winning rate as an indicator of the direction of the game. Deep_LSTM model uses advanced techniques including double Huber loss: the loss of two networks is calculated through a parallel structure. MAML predicting 3 games(Query)by using 28 games (support set)

Critically： We found that it is difficult for deep learning to handle the binary classification problem of Tabular Data, so we proposed Lasso-Ridge-based XGBoost.

## 7.2 Weaknesses

We believe that there is still room for improvement in the loss function. The definition of the loss function determines the upper limit of the model

The lack of interpretability of our model can be explained in the future by combining DeepLift model results

## Conclusion

In our research, we established three models, XGBoost, LSTM, and MAML, for modeling and analysis. We integrated the content of machine learning and deep learning. In the process of parameter optimization, we used grid search, gradient regression, regularization, and other methods to find the most suitable parameters for the model.

In terms of innovation, firstly, we used the Fibonacci sequence to quantify the momentum of athletes, and secondly, we used XGBoost ensemble learning based on elastic net regularization to predict the win rate of each game. In addition, we also applied the new deep learning concept of meta learning in Task 4 to model and complete transfer learning. Although our model also has certain shortcomings, for example, our XGBoost model has poor prediction results for individual matches, with an accuracy rate of only about 60% regardless of whether momentum is considered.

Overall, our model can lead us to the following conclusion: during a competition, the athlete's momentum often plays a decisive role in the outcome, especially the athlete's psychological momentum, which is often very significant. For evenly matched opponents, the role of strategic momentum may be more prominent.

After a visual process and indexing the results, we can accurately identify the time periods when athletes perform well or perform poorly. Afterwards, we conducted a controlled variable study on the feature of serving rounds, which has a decisive impact on the probability of players winning the game. Therefore, our model can confirm the view that the probability of serving winning the game/match point is much higher.

We successfully refuted the coach's statement using the XGBoost model. Our model found that momentum has a significant impact on the outcome of the competition. After incorporating the feature of momentum, the model's prediction of the winner's accuracy is very high, and the results are very optimistic.

We utilized the LSTM model and introduced the concept of gradient in the parameter optimization process. We differentiated the momentum we studied and then performed gradient regression, ultimately successfully analyzing the fluctuations in the game, that is, when the momentum will shift from leaning towards one player to leaning towards another.

We innovatively utilized the content of MAML meta learning to construct our model, aiming to achieve transfer learning. We use multiple Wimbledon datasets for training and then predict new matches. In addition, we also utilized data from other sports to test our model. We watched and recorded the men's singles final match between Fan Zhendong and Malone at the 2020 Tokyo Olympics frame by frame, and prepared the data according to the format of the Wimbledon dataset. Later, we used XGBoost to obtain the significant role of momentum in table tennis matches.

## Suggested methods

We divide momentum into two types: strategic momentum and psychological momentum. The role of strategic momentum: If the athlete's position differs from the competitor's, the athlete will generate asymmetric incentives, and "strategic momentum" will occur. Therefore, the winning athlete will put in more effort than the losing athlete. Mental momentum and past performance can change the perception of the contestant, which can have a causal effect on the contestant's subsequent performance. They affect the game's outcome to some extent, and from the model, psychological momentum significantly impacts the outcome.

In a tennis match, if there is a significant gap in the skill level of the two players, the athletes' play on the court will directly impact their psychological changes. Especially when dealing with the critical ball, if the athlete plays appropriately, it will lay an excellent psychological foundation for the next ball and subsequent games. If the handling is wrong, it will lead to the athlete's anxiety, impatience, and other destructive psychological emotions, hit self-confidence, and even affect the result of the game.

Furthermore, a reasonable and achievable competition goal is one of the essential means to enhance the confidence of athletes. In the meantime, self-motivation is an effective way to regulate oneself and prevent bad psychology by communicating with oneself. In the critical stage of the match or due to double errors and unforced errors, they can give themselves positive suggestions, such as "I can, I can do," or they can use "quick attention transfer" and other language directly related to the characteristics of tennis technology to implement self-suggestion, which will promote the positive psychological state of athletes, to improve the performance of athletes. Increase the likelihood of an athlete winning.

Through the above analysis, we can draw the following conclusions:
- Coaches should correctly use the time of public instruction to adjust the emotions and states of athletes quickly.
- Coaches need to help athletes set goals before each game
- Coaches ought to encourage athletes to engage in regular self-motivation.

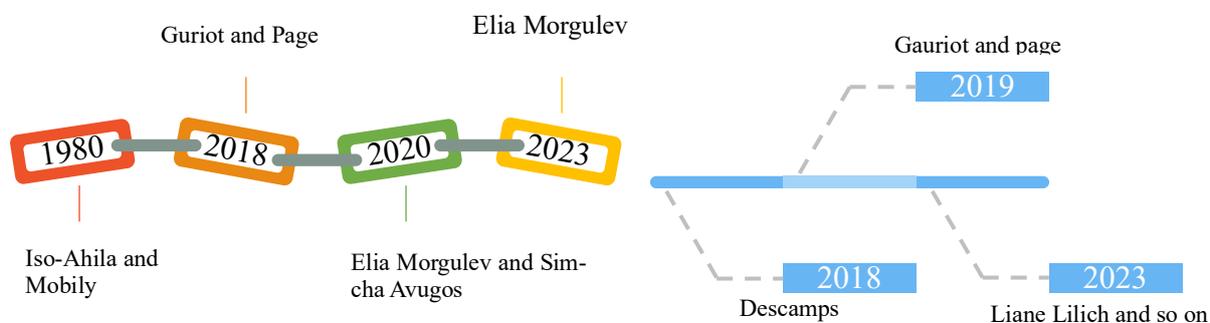

Figure 11　　The development of mental and strategic momentum